\documentclass[10pt,twocolumn,letterpaper]{article}
\usepackage{pbox}
\usepackage{cvpr}
\usepackage{times}
\usepackage{epsfig}
\usepackage{graphicx}
\usepackage{amsmath}
\usepackage{amssymb}

\usepackage[pagebackref=true,breaklinks=true,letterpaper=true,colorlinks,bookmarks=false]{hyperref}

\cvprfinalcopy 

\def\cvprPaperID{5052} 

\ifcvprfinal\fi
\begin{document}

\title{Cross-Modal Self-Attention Network for Referring Image Segmentation}

\author{Linwei Ye$^{\dagger}$ \qquad Mrigank Rochan$^{\dagger}$ \qquad Zhi Liu$^{\ddagger\ast}$ \qquad Yang Wang$^{\dagger}$\thanks{Zhi Liu and Yang Wang are the corresponding authors} \\
$^{\dagger}$University of Manitoba, Canada \quad $^{\ddagger}$Shanghai University, China \\
{\tt\small \{yel3, mrochan, ywang\}@cs.umanitoba.ca  \qquad liuzhi@staff.shu.edu.cn}
}

\maketitle
\thispagestyle{empty}

\begin{abstract}

We consider the problem of referring image segmentation. Given an input image and a natural language expression, the goal is to segment the object referred by the language expression in the image. Existing works in this area treat the language expression and the input image separately in their representations. They do not sufficiently capture long-range correlations between these two modalities. In this paper, we propose a cross-modal self-attention (CMSA) module that effectively captures the long-range dependencies between linguistic and visual features. Our model can adaptively focus on informative words in the referring expression and important regions in the input image. In addition, we propose a gated multi-level fusion module to selectively integrate self-attentive cross-modal features corresponding to different levels in the image. This module controls the information flow of features at different levels. We validate the proposed approach on four evaluation datasets. Our proposed approach consistently outperforms existing state-of-the-art methods. 
 
\end{abstract}

\section{Introduction}
\label{sec:intro}
Referring image segmentation is a challenging problem at the intersection of computer vision and natural language processing. Given an image and a natural language expression, the goal is to produce a segmentation mask in the image corresponding to entities referred by the the natural language expression~(see Fig.~\ref{fig:seg} for some examples). It is worth noting that the referring expression is not limited to specifying object categories (e.g. ``person'', ``cat''). It can take any free form language description which may contain appearance attributes (e.g. ``red'', ``long''), actions (e.g. ``standing'', ``hold'') and relative relationships (e.g. ``left'', ``above''), etc. Referring image segmentation can potentially be used in a wide range of applications, such as interactive photo editing and human-robot interaction. 

\begin{figure}[htb]
  \centering
  \includegraphics[width=8.2cm]{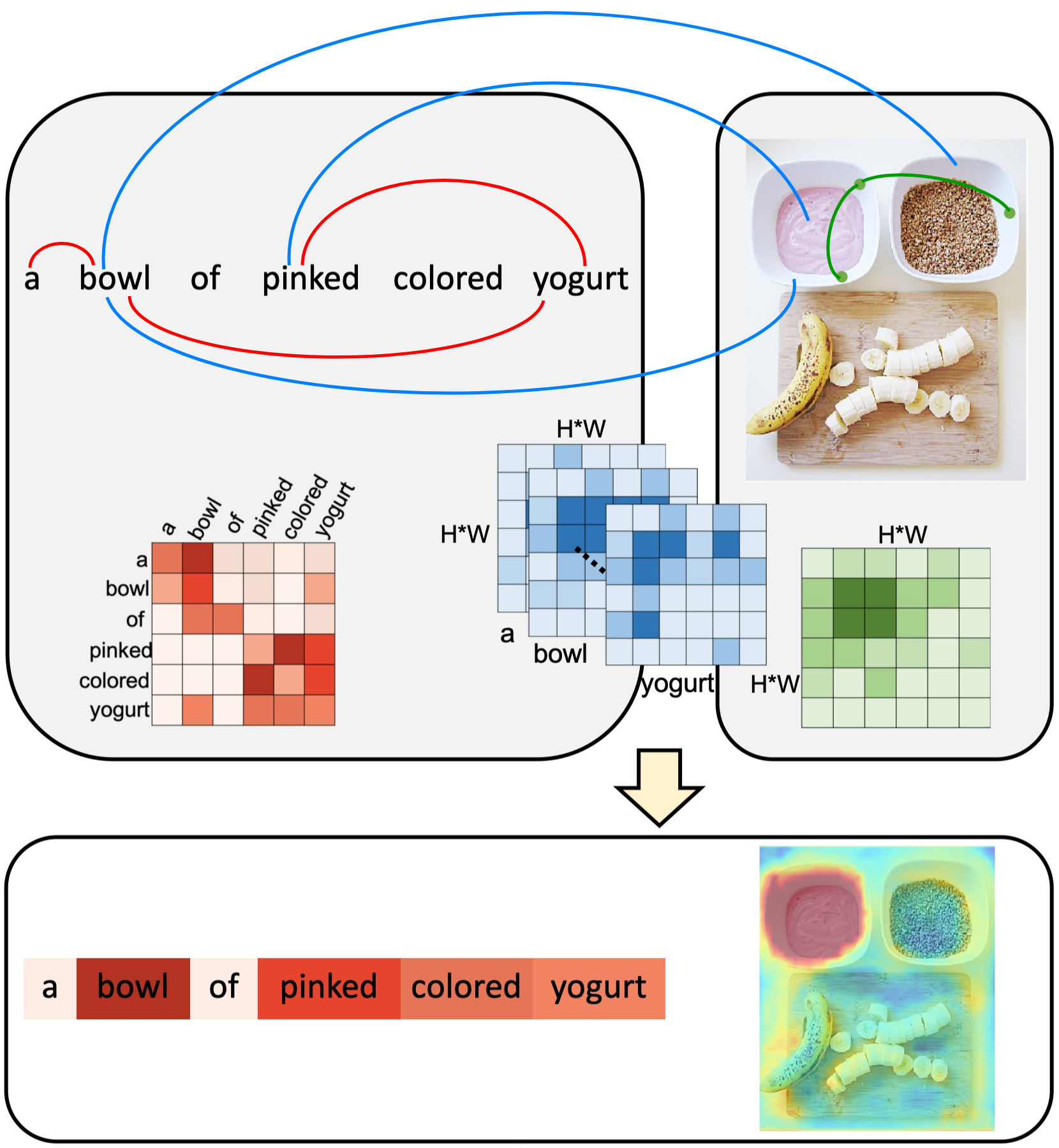}
\caption{(Best viewed in color) Illustration of our cross-modal self-attention mechanism. It is composed of three joint operations: self-attention over language (shown in red), self-attention over image representation (shown in green), and cross-modal attention between language and image (shown in blue). The visualizations of linguistic and spatial feature representations (in bottom row) show that the proposed model can focus on specific key words in the language and spatial regions in the image that are necessary to produce precise referring segmentation masks.} 
	
\label{fig:intro} 
\end{figure}

A popular approach (e.g. \cite{hu2016segmentation,li2018referring,shi2018key}) in this area is to use convolutional neural network (CNN) and recurrent neural network (RNN) to separately represent the image and the referring expression. The resultant image and language representations are then concatenated to produce the final pixel-wise segmentation result. The limitation of this approach is that the language encoding module may ignore some fine details of some individual words that are important to produce an accurate segmentation mask. 

Some previous works (e.g.~\cite{liu2017recurrent,margffoy2018dynamic}) focus on learning multimodal interaction in a sequential manner. The visual feature is sequentially merged with the output of LSTM-based~\cite{hochreiter1997long} language model at each step to infer a multimodal representation. However, the multimodal interaction only considers the linguistic and visual information individually within their local contexts. It may not sufficiently capture global interaction information essential for semantic understanding and segmentation. 

In this paper, we address the limitations of aforementioned methods. We propose a \textit{cross-modal self-attention (CMSA) module} to effectively learn long-range dependencies from multimodal features that represent both visual and linguistic information. Our model can adaptively focus on important regions in the image and informative keywords in the language description. Figure~\ref{fig:intro} shows an example that illustrates the cross-modal self-attention module, where the correlations among words in the language and regions in the image are presented. In addition, we propose a \textit{gated multi-level fusion module} to further refine the segmentation mask of the referred entity. The gated fusion module is designed to selectively leverage multi-level self-attentive features. 

In summary, this paper makes the following contributions: (1) A cross-modal self-attention method for referring image segmentation. Our model effectively captures the long-range dependencies between linguistic and visual contexts. As a result, it produces a robust multimodal feature representation for the task. (2) A gated multi-level fusion module to selectively integrate multi-level self-attentive features which effectively capture fine details for precise segmentation masks. (3) An extensive empirical study on four benchmark  datasets demonstrates that our proposed method achieves superior performance compared with state-of-the-art methods.


\section{Related Work}
\label{sec:related}

In this section, we review several lines of research related to our work  in the following fields.

\noindent\textbf{Semantic segmentation}: Semantic segmentation has achieved great advances in recent years. Fully convolutional networks (FCN)~\cite{long2015fully} take advantage of fully convolutional layers to train a segmentation model in an end-to-end way by replacing fully connected layers in CNN with convolutional layers. In order to alleviate the down-sampling issue and enlarge the semantic context, DeepLab~\cite{chen2016deeplab} adopts dilated convolution to enlarge the receptive field and uses atrous spatial pyramid pooling for multi-scale segmentation. An improved pyramid pooling module~\cite{zhao2017pyramid} further enhances the use of multi-scale structure. Lower level features are explored to bring more detailed information to complement high-level features for generating more accurate segmentation masks~\cite{badrinarayanan2015segnet,long2015fully,noh2015learning}.

\noindent\textbf{Referring image localization and segmentation}: In referring image localization, the goal is to localize specific objects in an image according to the description of a referring expression. It has been explored in natural language object retrieval~\cite{hu2016natural} and modelling relationship~\cite{hu2017modeling,yu2018mattnet}. In order to obtain a more precise result,  referring image segmentation is proposed to produce a segmentation mask instead of a bounding box. This problem was first introduced in \cite{hu2016segmentation}, where CNN and LSTM are used to extract visual and linguistic features, respectively. They are then concatenated together for spatial mask prediction. To better achieve word-to-image interaction, \cite{liu2017recurrent} directly combines visual features with each word feature from a language LSTM to recurrently refine segmentation results. Dynamic filter~\cite{margffoy2018dynamic} for each word further enhances this interaction. In~\cite{shi2018key}, word attention is incorporated in the image regions to model key-word-aware context. Low-level visual features are also exploited for this task in~\cite{li2018referring}, where Convolutional LSTM (ConvLSTM)~\cite{xingjian2015convolutional} progressively refines segmentation masks from high-level to low-level features sequentially. In this paper, we propose to adaptively integrate multi-level self-attentive features by the gated fusion module.

\noindent\textbf{Attention}: Attention mechanism has been shown to be a powerful technique in deep learning models and has been widely used in various tasks in natural language processing~\cite{bahdanau2014neural,vaswani2017attention} to capture keywords for context. In the multimodal tasks, word attention has been used to re-weight the importance of image regions for image caption generation~\cite{xu2015show}, image question answering~\cite{yang2016stacked} and referring image segmentation~\cite{shi2018key}. In addition, attention is also used for modeling subject, relationship and object~\cite{hu2017modeling} and for referring relationship comprehension~\cite{yu2018mattnet}. The diverse attentions of query, image and objects are calculated separately and then accumulated circularly for visual grounding in~\cite{deng2018visual}.

Self-attention~\cite{vaswani2017attention} is proposed to attend a word to all other words for learning relations in the input sequence. It significantly improves the performance for machine translation. This technique is also introduced in videos to capture long-term dependencies across temporal frames~\cite{wang2018non}. Different from these works, we propose a cross-modal self-attention module to bridge attentions across language and vision.

\begin{figure*}[ht]
\begin{center}
\includegraphics[width=0.93\textwidth]{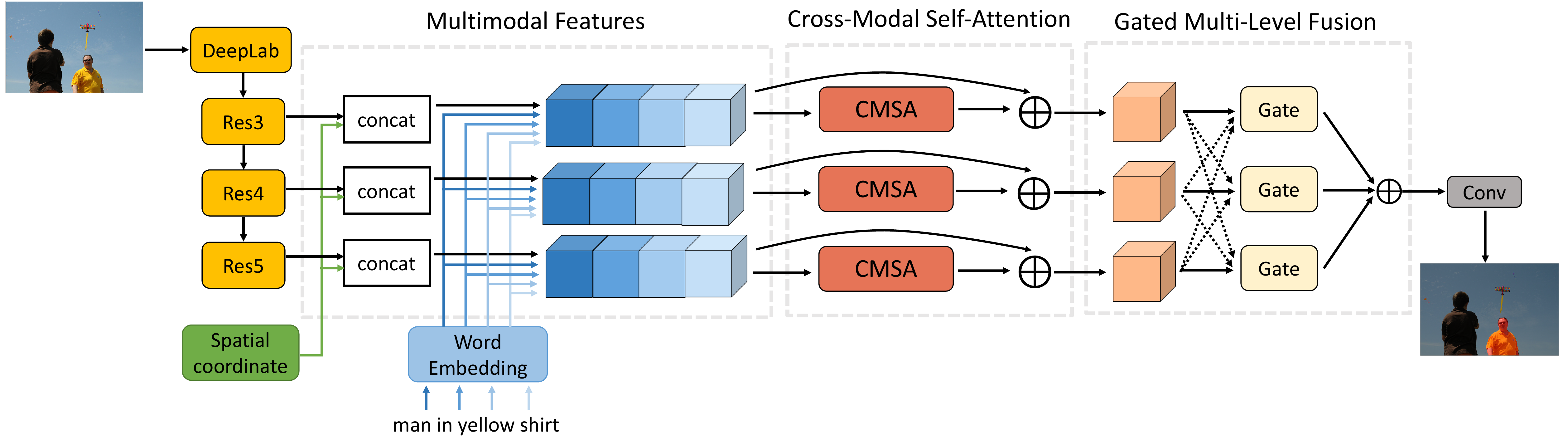}
\end{center}
   \caption{An overview of our approach. The proposed model consists of three components including multimodal features, cross-modal self-attention (CMSA) and a gated multi-level fusion. Multimodal features are constructed from the image feature, the spatial coordinate feature and the language feature for each word. Then the multimodual feature at each level is fed to a cross-modal self-attention module to build long-range dependencies across individual words and spatial regions. Finally, the gated multi-level fusion module combines the features from different levels to produce the final segmentation mask.}
\label{fig:overview}
\end{figure*}

\section{Our Model}\label{sec:approach}

The overall architecture of our model is shown in Fig.~\ref{fig:overview}. Given an image and a referring expression as the query, we first use a CNN to extract visual feature maps at different levels from the input image. Each word in the referring expression is represented as a vector of word embedding. Every word vector is then appended to the visual feature map to produce a multimodal feature map. Thus, there is a multimodal feature map for each word in the referring expression. We then introduce self-attention~\cite{vaswani2017attention} mechanism to combine the feature maps of different words into a cross-modal self-attentive feature map. The self-attentive feature map captures rich information and long-range dependencies of both linguistic and visual information of the inputs. In the end, the self-attentive features from multiple levels are combined via a gating mechanism to produce the final features used for generating the segmentation output.

Our model is motivated by several observations. First of all, in order to solve referring image segmentation, we typically require detailed information of certain individual words (e.g. words like ``left'', ``right''). Previous works~(e.g.~\cite{hu2016segmentation,li2018referring,shi2018key}) take word vectors as inputs and use LSTM to produce a vector representation of the entire referring expression. The vector representation of the entire  referring expression is then combined with the visual features for referring image segmentation. The potential limitation of this technique is that the vector representation produced by LSTM captures the meaning of the entire referring expression while missing sufficiently detailed information of some individual words needed for the referring image segmentation task. Our model addresses this issue and does not use LSTM to encode the entire referring expression. Therefore, it can better capture more detailed word-level information. Secondly, some previous works~(e.g.~\cite{liu2017recurrent,margffoy2018dynamic}) process each word in the referring expression and concatenate it with visual features to infer the referred object in a sequential order using a recurrent network. The limitation is that these methods only look at local spatial regions and lack the interaction over long-range spatial regions in global context which is essential for semantic understanding and segmentation. In contrast, our model uses a cross-modal self-attention module that can effectively model long-range dependencies between linguistic and visual modalities. Lastly, different from~\cite{li2018referring} which adopts ConvLSTM to refine segmentation with multi-scale visual features sequentially, the proposed method employs a novel gated fusion module for combining multi-level self-attentive features.

\subsection{Multimodal Features}
The input to our model consists of an image $I$ and a referring expression with $N$ words $w_n, n\in{1, 2,...,N}$. We first use a backbone CNN network to extract visual features from the input image. The feature map extracted from a specific CNN layer is represented as $V\in\mathbb{R}^{H\times W\times C_v}$, where $H$, $W$ and $C_v$ are the dimensions of height, width and feature channel, respectively. For ease of presentation, we only use features extracted from one particular CNN layer for now. Later in Sec.~\ref{sec:gf}, we present an extension of our method that uses features from multiple CNN layers. 

For the language description with $N$ words, we encode each word $w_n$ as a one-hot vector, and project it into a compact word embedding represented as $\mathbf{e}_n\in\mathbb{R}^{C_l}$ by a lookup table. Different from previous methods~\cite{hu2016segmentation,li2018referring,shi2018key} that apply LSTM to process the word vectors sequentially and encode the entire language description as a sentence vector, we keep the individual word vectors and introduce a cross-modal self-attention module to capture long-range correlations between these words and spatial regions in the image. More details will be presented in Sec.~\ref{sec:msa}.


In addition to visual features and word vectors, spatial coordinate features have also been shown to be useful for referring image segmentation~\cite{hu2016segmentation,li2018referring,liu2017recurrent}. Following prior works, we define an $8$-D spatial coordinate feature at each spatial position using the implementation in~\cite{liu2017recurrent}. The first $3$-dimensions of the feature map encode the normalized horizontal positions. The next $3$-dimensions encode normalized vertical positions. The last $2$-dimensions encode the normalized width and height information of the image.

Finally, we construct a joint multimodal feature representation at each spatial position for each word by concatenating the visual features, word vectors, and spatial coordinate features. Let $p$ be a spatial location in the feature map $V$, i.e. $p\in\{1,2,..., H\times W\}$. We use $\mathbf{v}_p\in\mathbb{R}^{C_v}$ to denote the ``slice'' of the visual feature vector at the spatial location $p$. The spatial coordinate feature of the location $p$ is denoted as $\mathbf{s}_p\in\mathbb{R}^8$. Thus we can define the multimodal feature $\mathbf{f}_{pn}$ corresponding to the location $p$ and the $n$-th word as follows:
\begin{equation}
\mathbf{f}_{pn} = Concat\left(\frac{\mathbf{v}_p}{||\mathbf{v}_p||_2}, \frac{\mathbf{e}_n}{||\mathbf{e}_n||_2}, \mathbf{s}_p\right)
\end{equation}
where $||\cdot||_2$ denotes the $L_2$ norm of a vector and $Concat(\cdot)$ denotes the concatenation of several input vectors. The feature vector $\mathbf{f}_{pn}$ encodes information about the combination of a specific location $p$ in the image and the $n$-th word $w_{n}$ in the referring expression with a total dimension of $(C_v+C_l+8)$. We use $F=\{\mathbf{f}_{pn}: \forall p, \forall n\}$ to represent the collection of features $\mathbf{f}_{pn}$ for different spatial locations and words. The dimension of $F$ is $N\times H\times W\times (C_v+C_l+8)$.


\subsection{Cross-Modal Self-Attention}\label{sec:msa}
The multimodal feature $F$ is quite large which may contain a lot of redundant information. Additionally, the size of $F$ is variable depending on the number of words in the language description. It is difficult to directly exploit $F$ to produce the segmentation output. In recent years, the attention mechanism~\cite{hu2017modeling,shi2018key,vaswani2017attention,xu2015show,yu2018mattnet} has been shown to be a powerful technique that can capture important information from raw features in either linguistic or visual representation. Different from above works, we propose a cross-modal self-attention module to jointly exploit attentions over multimodal features. In particular, inspired by the success of self-attention~\cite{vaswani2017attention,wang2018non}, the designed cross-modal self-attention module can capture long-range dependencies between the words in a referring expression and different spatial locations in the input image. The proposed module takes $F$ as the input and produces a feature map that summarizes $F$ after learning the correlation between the language expression and the visual context. Note that the size of this output feature map does not depend on the number of words present in the language description.



Given a multimodal feature vector $\mathbf{f}_{pn}$, the cross-modal self-attention module first produces a set of query, key and value pair by linear transformations as $\mathbf{q}_{pn}=W_q\mathbf{f}_{pn}$ , $\mathbf{k}_{pn}=W_k\mathbf{f}_{pn}$ and  $\mathbf{v}_{pn}=W_v\mathbf{f}_{pn}$ at each spatial location $p$ and the $n$-th word, where $\{W_q, W_k, W_v\}$ are part of the model parameters to be learned. Each query, key and value is reduced from the high dimension of multimodal features to the dimension of $512$ in our implementation, i.e. $W_q,W_k,W_v\in\mathbb{R}^{512\times (C_v+C_l+8)}$, for computation efficiency.

We compute the cross-modal self-attentive feature $\hat{v}_{pn}$ as follows: 
\begin{eqnarray}
  && \hspace{-20pt} \widehat{\mathbf{v}}_{pn} = \sum_{p'}\sum_{n'}a_{p,n,p',n'} \mathbf{v}_{p'n'}, \ \ \ \textrm{where }\label{eq:self1}\\
  && \hspace{-20pt} a_{p,n,p',n'}=\textrm{Softmax}(\mathbf{q}_{p'n'}^T \mathbf{k}_{pn})\label{eq:self2}
\end{eqnarray}
where $a_{p,n,p',n'}$ is the attention score that takes into account of the correlation between $(p,n)$ and any other combinations of spatial location and word $(p',n')$.

Then $\widehat{\mathbf{v}}_{pn}$ is transformed back to the same dimension as $\mathbf{f}_{pn}$ via a linear layer and is added element-wise with $\mathbf{f}_{pn}$ to form a residual connection. This allows the insertion of this module into to the backbone network without breaking its  behavior~\cite{he2016deep}. The final feature representation is average-pooled over all words in the referring expression. These operations can be summarized as:
\begin{equation}
    \widehat{\mathbf{f}}_p =  \textrm{avg-pool}_n(W_{\widehat{v}}\widehat{\mathbf{v}}_{pn}+ \mathbf{f}_{pn} ) = \frac{\sum_{n=1}^N \left(W_{\widehat{v}}\widehat{\mathbf{v}}_{pn}+ \mathbf{f}_{pn}\right)}{N}\label{eq:final}
\end{equation}
where $W_{\widehat{v}}\in\mathbb{R}^{(C_v+C_l+8)\times 512}$ and $\widehat{\mathbf{f}}_p\in\mathbb{R}^{C_v+C_l+8}$. We use $\widehat{F}=\{\widehat{\mathbf{f}}_p: \forall p\}$ to denote the collection of $\widehat{\mathbf{f}}_{p}$ at all spatial locations, i.e. $\widehat{F}\in\mathbb{R}^{H\times W\times (C_v+C_l+8)}$.

\begin{figure}[htb]
  \centering
  \includegraphics[width=8.3cm]{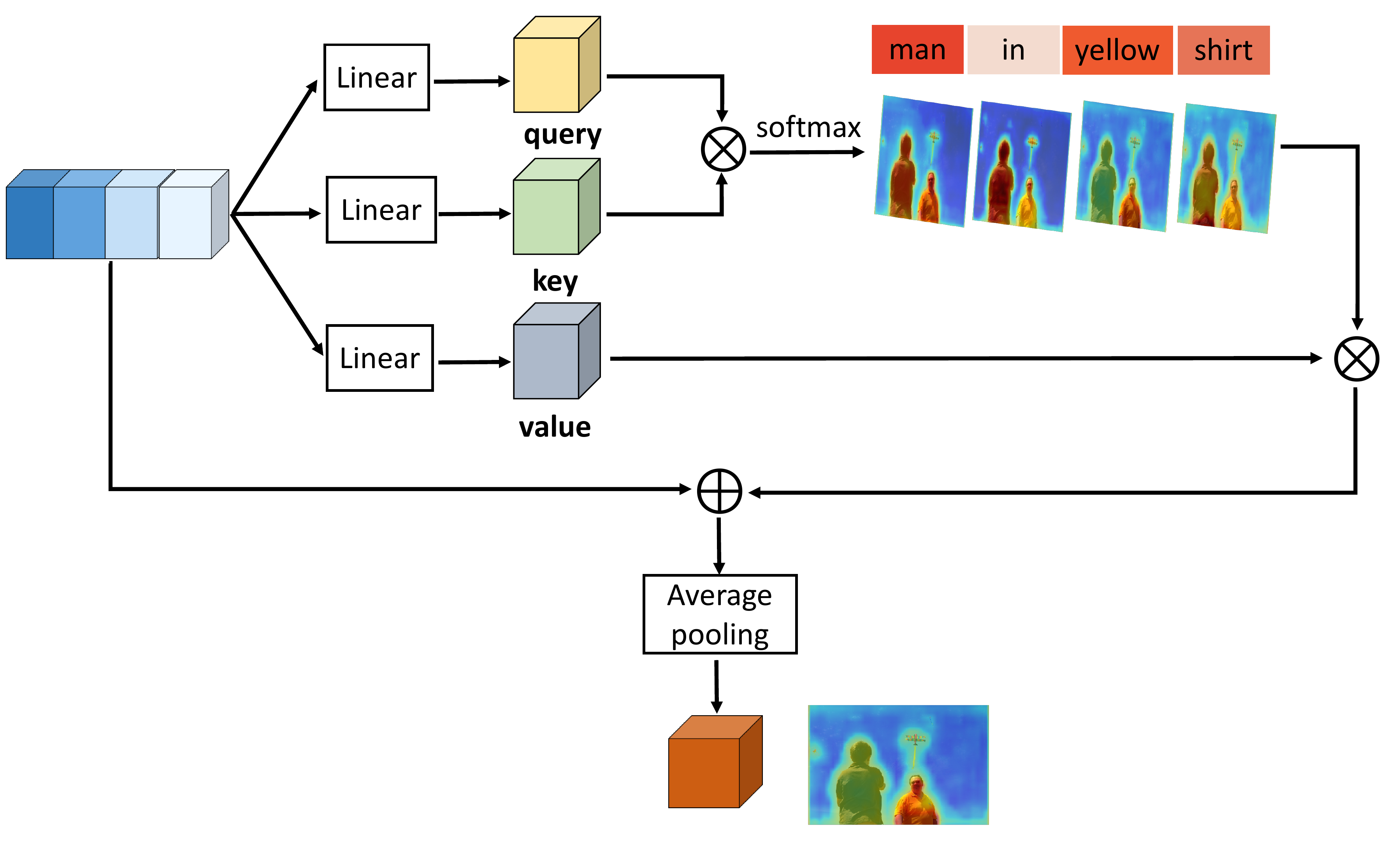}
    \caption{An illustration of the process of generating the cross-modal self-attentive (CMSA) feature from an image and a language expression (``man in yellow shirt''). We use $\otimes$ and $\oplus$ to denote matrix multiplication and element-wise summation, respectively. The softmax operation is performed over each row which indicates the attentions across each visual and language cell in the multimodal feature. We visualize the internal linguistic and spatial representations. Please refer to Sec.~\ref{sec:result} and Sec.~\ref{sec:vis} for more details.}
\label{fig:nl}
\end{figure}

Figure~\ref{fig:nl} illustrates the process of generating cross-modal self-attentive features. 

\subsection{Gated Multi-Level Fusion}\label{sec:gf}
The feature representation $\widehat{F}$ obtained from Eq.~\ref{eq:final} is specific to a particular layer in CNN. Previous work~\cite{li2018referring} has shown that fusing features at multiple scales can improve the performance of referring image segmentation. In this section, we introduce a novel gated fusion technique to integrate multi-level features.

Let $\widehat{F^{(i)}}$ be the cross-modal self-attentive feature map at the $i$-th level. Following~\cite{li2018referring}, we use ResNet based DeepLab-101 as the backbone CNN and consider feature maps at three levels ($i=1,2,3$) corresponding to ResNet blocks $Res3$, $Res4$ and $Res5$. Let $C_{v_{i}}$ be the channel dimension of the visual feature map at the $i$-th level of the network. We use $\widehat{F^{(i)}}=\{\widehat{\mathbf{f}}^{(i)}_{p}: \forall p\}$ to indicate the collection of cross-modal self-attentive features $\widehat{\mathbf{f}}^{(i)}_{p}\in\mathbb{R}^{C_{v_{i}}+C_l+8}$ for different spatial locations corresponding to the $i$-th level. Our goal is to fuse the feature maps $\widehat{F^{(i)}}$ ($i=1,2,3$) to produce a fused feature map for producing the final segmentation output.

Note that the feature maps $\widehat{F^{(i)}}$  have different channel dimensions at different level $i$. At each level, we apply a $1\times 1$ convolutional layer to make the channel dimensions of different levels consistent and result in an output $X^{(i)}$.

For the $i$-th level, we generate a memory gate $m^{i}$ and a reset gate $r^{i}$ ($r^{i}, m^{i}\in\mathbb{R}^{H_i\times W_i}$), respectively. These gates play a similar role to the gates in LSTM. Different from stage-wise memory updates~\cite{cho2014learning,hochreiter1997long}, the computation of gates at each level is decoupled from other levels. The gates at each level control how much the visual feature at each level contributes to the final fused feature. Each level also has a contextual controller $G^{i}$ which modulates the information flow from other levels to the $i$-th level. This process can be summarized as: 
\begin{align}
\label{eq:mgi}
\begin{split}
\hspace{-10pt}    G^{i} & = (1 - m^{i}) \odot X^{i} + \sum_{j\in \{1,2,3\} \setminus \{i\}}\gamma^{j} m^{j} \odot X^{j}\\
\hspace{-10pt}    F^{i}_o & = r^{i} \odot \textrm{tanh}(G^{i}) + (1-r^{i}) \odot X^i, \ \ \forall i \in \{1,2,3\}\\
\end{split}
\end{align}
where $\odot$ denotes Hadamard product. $\gamma^j$ is a learnable parameter to adjust the relative ratio of the memory gate which controls information flow of features from different levels $j$ combined to the current level $i$. 

In order to obtain the segmentation mask, we aggregate the feature maps $F^i_{o}$ from the three levels and apply a $3\times 3$ convolutional layer followed by the sigmoid function. This sequence of operations outputs a probability map ($P$) indicating the likelihood of each pixel being the foreground in the segmentation mask, i.e.:
\begin{equation}
P = \sigma\left(\mathbb{C}_{3\times 3} \left(\sum_{i=1}^{3} F_{o}^i\right)\right)
\end{equation}
where $\sigma(\cdot)$ and $\mathbb{C}_{3\times 3}$ denote the sigmoid and $3\times 3$ convolution operation, respectively. A binary cross-entropy loss function is defined on the predicted output and the ground-truth segmentation mask $Y$ as follows:
\begin{equation}
    L = \frac{1}{\Omega}\sum_{m=1}^{\Omega}(Y(m)\log P(m)+(1-Y(m))\log(1-P(m)))\label{eq:final_loss}
\end{equation}
where $\Omega$ is the whole set of pixels in the image and $m$ is m-th pixel in it. We use the Adam algorithm~\cite{kingma2014adam} to optimize the loss in Eq.~\ref{eq:final_loss}.



\begin{table*}[ht!]
\begin{center}
\begin{tabular}{|c|ccc|ccc|c|c|}
\hline
	 &  &UNC& & &UNC+& & G-Ref  & ReferIt\\

	    & val &  testA &  testB &  val &  testA &  testB & val  & test\\
\hline
LSTM-CNN~\cite{hu2016segmentation} & - &  - &  -  		    &   - &  - &  -  			& 28.14 & 48.03 \\
RMI~\cite{liu2017recurrent} & 45.18 &  45.69 &  45.57 & 29.86 &  30.48 &  29.50  & 34.52 & 58.73 \\
DMN~\cite{margffoy2018dynamic}  & 49.78 &  54.83 &  45.13 & 38.88 &  44.22 &  32.29  & 36.76 & 52.81 \\
KWA~\cite{shi2018key}  & - &  - &  -  & - &  - &  -  & 36.92 & 59.09 \\
RRN~\cite{li2018referring}      & 55.33 &  57.26 &  53.93 & 39.75 &  42.15 &  36.11  & 36.45 & 63.63 \\
Ours		& \textbf{58.32} &  \textbf{60.61} &  \textbf{55.09}      &   \textbf{43.76} &  \textbf{47.60} &  \textbf{37.89}         & \textbf{39.98} & \textbf{63.80}  \\
\hline
\end{tabular}
\end{center}
\caption{Comparison of segmentation performance with the state-of-the-art methods on four evaluation datasets in terms of IoU.}
\label{tab:miou}
\end{table*}

\begin{table}[ht!]
\begin{center}
\begin{tabular}{|c|c|}
\hline
Method  & IoU \\
\hline  
No attention &  45.63    \\
Word attention  & 47.01  \\
Pixel attention   &   47.84 \\ 
Word-pixel pair attention & 47.57  \\
Cross-modal self-attention & {\bf 50.12} \\
\hline
\end{tabular}
\end{center}
\caption{Ablation study of different attention methods for multimodal features on the UNC val set.}
\label{tab:ab_att}
\end{table}
\begin{table*}[ht!]
\begin{center}
\begin{tabular}{|c|c|c|c|c|c|c|}
\hline
    Method &  prec@0.5 & prec@0.6 &  prec@0.7  & prec@0.8 &  prec@0.9 & IoU\\
\hline
    RMI-LSTM~\cite{liu2017recurrent} & 42.99  & 33.24 & 22.75 & 12.11 & 2.23& 45.18   \\
    RRN-CNN~\cite{li2018referring}$^*$  &  47.59 & 38.76 & 26.53 & 14.79 & 3.17 & 46.95   \\
    CMSA-S  & 51.19  & 41.31 & 29.57 & 14.99 & 2.61 & 48.53   \\
    CMSA-W  & \bf{51.95}  &\bf{43.11} & \bf{32.74} & \bf{19.28} & \bf{4.11} & \bf{50.12}   \\
\hline
    CMSA+PPM  & 58.25  & 49.82 & 39.09 & 24.76 & 5.73 & 53.54   \\
    CMSA+Deconv  & 58.29  & 49.94 & 39.16 & 25.42 & 6.75 & 54.18 \\
    CMSA+ConvLSTM  & 64.73  & 56.03 & 45.23 & 29.15 & 7.86 & 56.56   \\
    CMSA+Gated  & 65.17  & 57.25 & 47.37 & 33.31 & 9.66 &  57.08  \\ 
    CMSA+GF(Ours)  & \bf{66.44}  & \bf{59.70} & \bf{50.77} & \bf{35.52} & \bf{10.96} & \bf{58.32}  \\
    
\hline
\end{tabular}
\end{center}
\caption{Ablation study on the UNC val set. The top four methods compare results of different methods for multimodal feature representations. The bottom five results show comparisons of multi-level feature fusion methods. CMSA and GF denote the proposed cross-modal self-attention and gated multi-level fusion modules. All methods use the same base model (DeepLab-101) and DenseCRF for postprocessing. $^*$The numbers for~\cite{li2018referring} are slightly higher than original numbers reported in their paper which did not use DenseCRF postprocessing.}
\label{tab:ab}
\end{table*}

\section{Experiments}\label{sec:experiment}
In this section, we first introduce the datasets and experimental setup in Sec.~\ref{sec:dataset}. Then we present the main results of our method and compare with other state-of-the-art in Sec.~\ref{sec:result}. Finally, we perform detailed ablation analysis to demonstrate the relative contribution of each component of our proposed method in Sec.~\ref{sec:ab}. We also provide  visualization and failure cases to help gain insights of our model in Sec.~\ref{sec:vis}.

\subsection{Datasets and Setup}\label{sec:dataset}

\noindent{\bf Implementation details}: 
Following previous work~\cite{li2018referring,liu2017recurrent, shi2018key}, we keep the maximum length of query expression as $20$ and embed each word to a vector of $C_l=1000$ dimensions. Given an input image, we resize it to $320\times320$ and use the outputs of DeepLab-101 ResNet blocks $Res3,Res4,Res5$ as the inputs for multimodal features. The dimension used in $X^{(i)}$ for gated fusion is fixed to $500$.
The network is trained with 
an initial learning rate of $2.5e^{-4}$ and weight decay of $5e^{-4}$. The learning rate is gradually decreased using the polynomial policy with power of $0.9$. For fair comparisons, the final segmentation results are refined by DenseCRF~\cite{krahenbuhl2011efficient}.

\noindent{\bf Datasets}: We perform extensive experiments on four referring image segmentation datasets: UNC~\cite{yu2016modeling}, UNC+~\cite{yu2016modeling}, G-Ref~\cite{mao2016generation} and ReferIt~\cite{kazemzadeh2014referitgame}.

The UNC dataset contains 19,994 images with 142,209 referring expressions for 50,000 objects. All images and expressions are collected from the MS COCO~\cite{lin2014microsoft} dataset interactively with a two-player game~\cite{kazemzadeh2014referitgame}. Two or more objects of the same object category appear in each image. 

The UNC+ dataset is similar to the UNC dataset. but with a restriction that no location words are allowed in the referring expression. In this case, expressions regarding referred objects totally depend on the appearance and the scene context. It consists of 141,564 expressions for 49,856 objects in 19,992 images.

The G-Ref dataset is also collected based on MS COCO. It contains of 104,560 expressions referring to 54,822 objects from 26,711 images. Annotations of this dataset come from Amazon Mechanical Turk instead of a two-player game. The average length of expressions is 8.4 words which is longer than that of other datasets (less than 4 words). 

The ReferIt dataset is built upon the IAPR TC-12~\cite{escalante2010segmented} dataset. It has 130,525 expressions referring to 96,654 distinct object masks in 19,894 natural images. In addition to objects, it also contains annotations for stuff classes such as water, sky and ground.

\noindent{\bf Evaluation metrics:} Following previous work~\cite{li2018referring,liu2017recurrent, shi2018key}, we use intersection-over-union $(IoU)$ and $prec@X$ as the evaluation metrics. The $IoU$ metric is a ratio between intersection and union of the predicted segmentation mask and the ground truth. The $prec@X$ metric measures the percentage of test images with an IoU score higher than the threshold $X$, where $X \in\{ 0.5, 0.6,0.7,0.8,0.9\}$ in the experiments.

\begin{figure*}[!htb]
  \centering
{\setlength{\tabcolsep}{0.5pt}
\scalebox{0.92}{
  \begin{tabular}{ccccc}


           \multicolumn{5}{c}{Query:  \textit{``the bottom two luggage cases being rolled''}}    \\
    \includegraphics[width=1.2in]{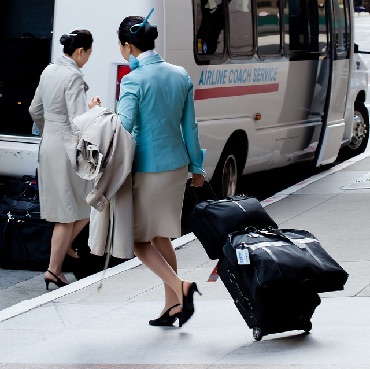}&
    \includegraphics[height=0.9in]{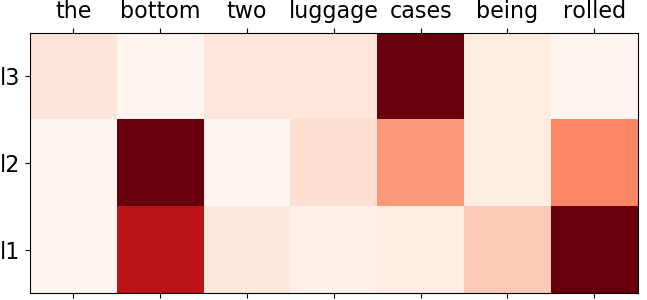}&
    \includegraphics[width=1.2in]{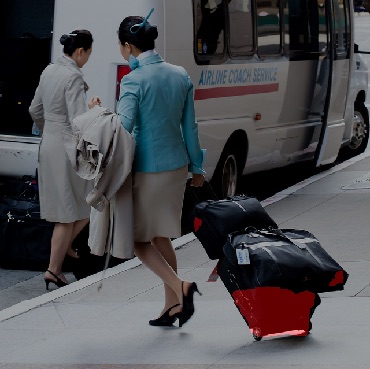}&
     \includegraphics[width=1.2in]{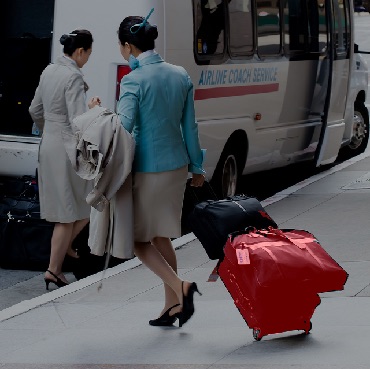}&
      \includegraphics[width=1.2in]{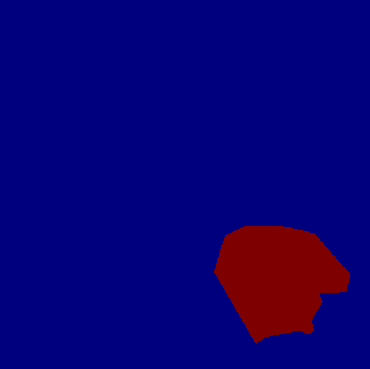}\\

    \multicolumn{5}{c}{Query:  \textit{``small black suitcase''}} \\
    \includegraphics[width=1.2in]{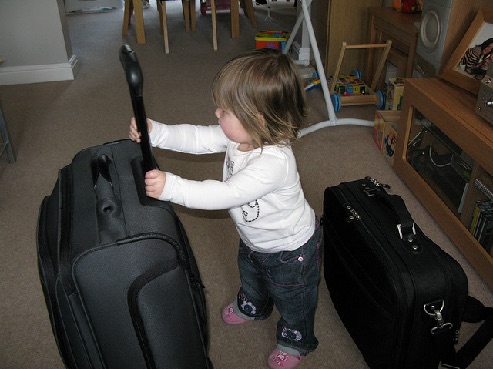}&
    \includegraphics[height=0.9in]{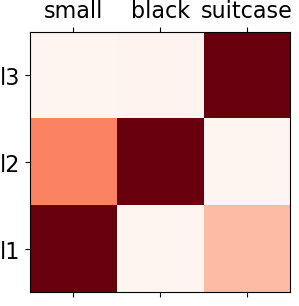}&
    \includegraphics[width=1.2in]{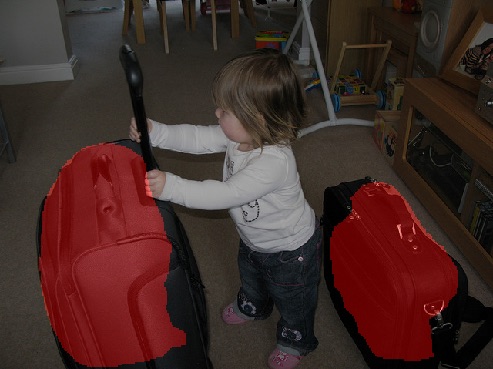}&
     \includegraphics[width=1.2in]{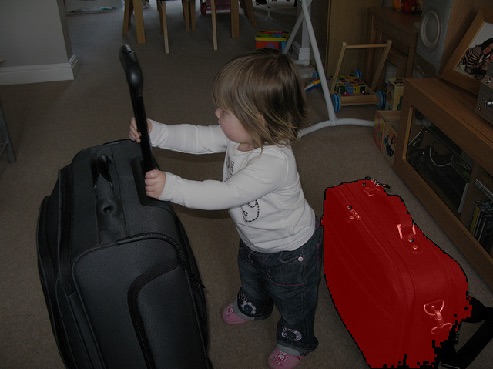}&
      \includegraphics[width=1.2in]{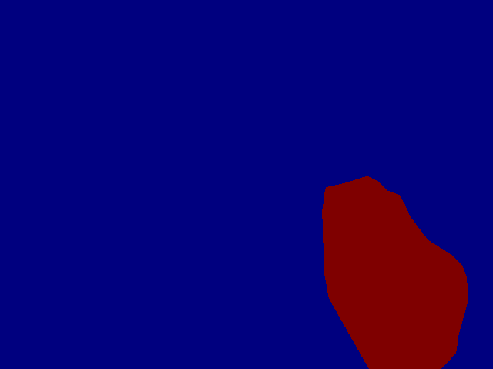}\\
        
      
    \multicolumn{5}{c}{Query:  \textit{``second vase from right''}} \\
    \includegraphics[width=1.2in]{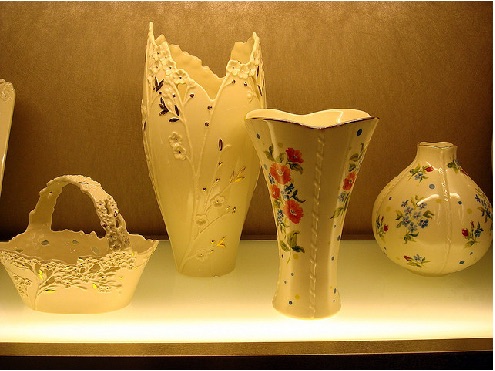}&
    \includegraphics[height=0.9in]{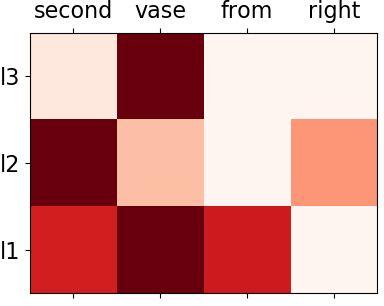}&
    \includegraphics[width=1.2in]{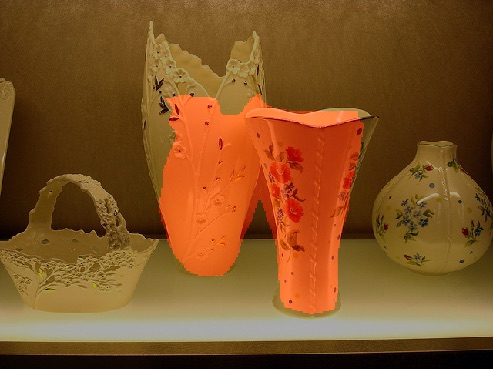}&
     \includegraphics[width=1.2in]{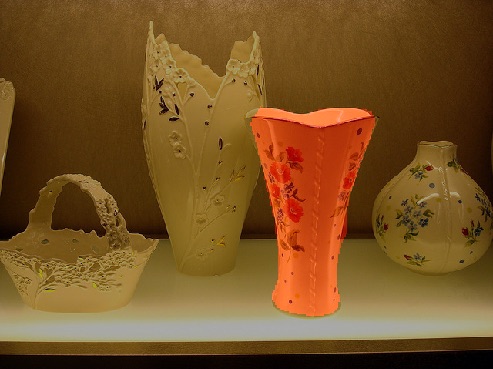}&
      \includegraphics[width=1.2in]{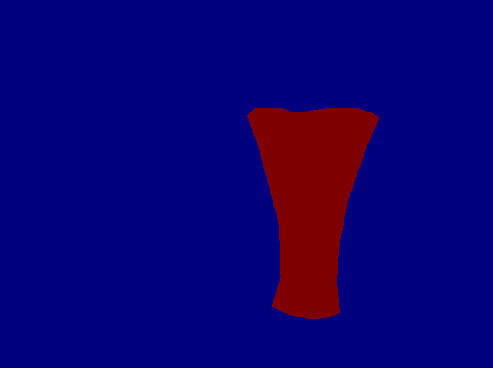}\\     
        
        (a) & (b) & (c) &(d)&(e)\\
  \end{tabular}
}
}
\caption{Qualitative examples of referring image segmentation: (a) original image; (b) visualization of the linguistic representation (attentions to word at each of the three feature levels); (c) segmentation result using only features at the 3rd level (i.e. $Res5$); (d) segmentation result using multi-level features and; (e) ground truth.}
\label{fig:seg}
\end{figure*}

\subsection{Experimental Evaluation}\label{sec:result}
\noindent{\bf Quantitative results:} Table~\ref{tab:miou} presents comparisons of our method with existing state-of-the-art approaches. Our proposed method consistently outperforms all other methods  on all four datasets. The improvement is particularly significant on the more challenging datasets, such as UNC+ which has no location words and G-Ref which contains longer and richer query expressions. This demonstrates the advantage of capturing long-range dependencies for cross-modal features and capturing the referred objects based on expressions by our model. 

\noindent{\bf Qualitative results:} Figure~\ref{fig:seg} shows some qualitative examples generated by our network. To better understand the benefit of multi-level self-attentive features, we visualize the linguistic representation to show attention distributions at different levels. For a given level, we get the collection of the attention scores $\{a_{p,n,p',n'}: \forall p, \forall n, \forall p', \forall n'\}$ in Eq.~\ref{eq:self2} and average over the dimensions $p$, $p'$ and $n'$. Thereby we can get a vector of length $N$. We repeat this operation for all three levels and finally obtain a matrix of $3\times N$. This matrix is shown in Fig.~\ref{fig:seg} (2nd column). We can see that the attention distribution over words corresponding to a particular feature level is different. Features at higher levels (e.g. $l_3$) tend to focus on words that refer to objects (e.g. ``suitcase'', ``vase''). Features at lower levels (e.g. $l_1$, $l_2$) tend to focus on words that refer to attributes (e.g. ``black'') or relationships (e.g. ``bottom'', ``second'').

\subsection{Ablation Study}\label{sec:ab}

We perform additional ablation experiments on the UNC dataset to further investigate the relative contribution of each component of our proposed model.

\noindent{\bf Attention methods}: We  first perform experiments on different attention methods for multimodal features. We alternatively use no attention, word attention, pixel attention and word-pixel pair attention by zeroing out the respective components in Eq.~\ref{eq:self1}. As shown in Table~\ref{tab:ab_att}, the proposed cross-modal self-attention outperforms all other attention methods significantly. This demonstrates the language-to-vision correlation can be better learned together within our cross-modal self-attention method.

\noindent{\bf Multimodal feature representation}: This experiment evaluates the effectiveness of the multimdoal feature representation. Similar to the baselines, i.e. multimodal LSTM interaction in~\cite{liu2017recurrent} and convolution integration in~\cite{li2018referring}, we directly take the output of the $Res5$ of the network to test the performance of multimodal feature representation without the multi-level fusion. We use CMSA-W to denote the proposed method in Sec.~\ref{sec:msa}. In addition, a variant method CMSA-S which also uses the same cross-modal self-attentive feature, instead encodes the whole sentence to one single language vector by LSTM.

As shown in Table~\ref{tab:ab} (top 4 rows), the proposed cross-modal self-attentive feature based approaches achieve significantly better performance than other baselines. Moreover, the word based method CMSA-W outperforms sentence based method CMSA-S for multimodal feature representation.

\noindent{\bf Multi-level feature fusion}: This experiment verifies the relative contribution of the proposed gated multi-level fusion module. Here we use our cross-modal self-attentive features as inputs and compare with several well-known feature fusion techniques, such as Deconv~\cite{noh2015learning} and PPM~\cite{zhao2017pyramid} in semantic segmentation and ConvLSTM~\cite{li2018referring} in referring image segmentation. 

In order to clearly understand the benefit of our fusion method, we also develop another self-gated method that uses the same gate generation method in Sec.~\ref{sec:gf} to generate memory gates and directly multiply by its own features without interactions with features from other levels. As presented in the bottom 5 rows in Table~\ref{tab:ab}, the proposed gated multi-level fusion outperforms these other multi-scale feature fusion methods.

\begin{figure*}[!htb]
  \centering
{\setlength{\tabcolsep}{0.5pt}
\scalebox{0.9}{
  \begin{tabular}{ccccc}

    \includegraphics[width=1.21in]{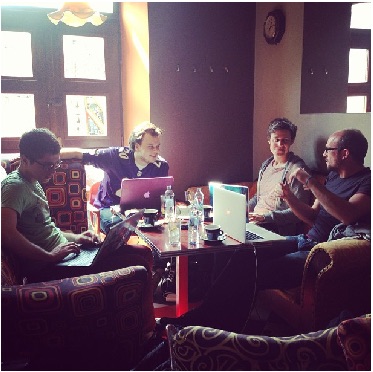}&
    \includegraphics[width=1.21in]{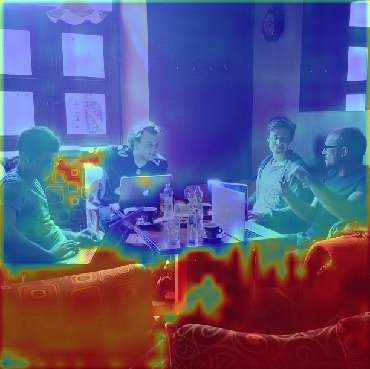}&
    \includegraphics[width=1.21in]{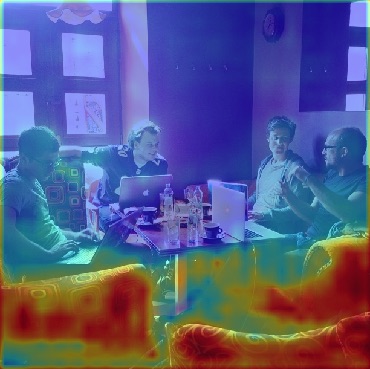}& 
    \includegraphics[width=1.21in]{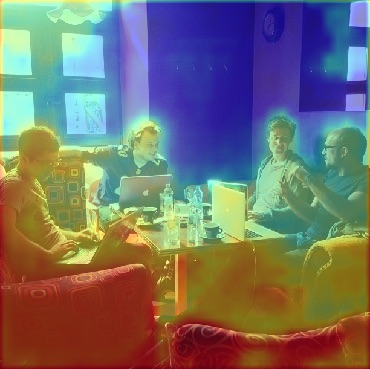}&
     \includegraphics[width=1.21in]{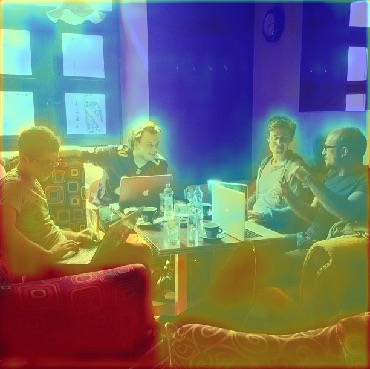}\\
     
     & \textit{``chair"} & \textit{``couch"} & \textit{``partial chair in front"} & \textit{``bottom left portion}\\ &&&&  \textit{of couch on left"} \\
   
       \includegraphics[width=1.21in]{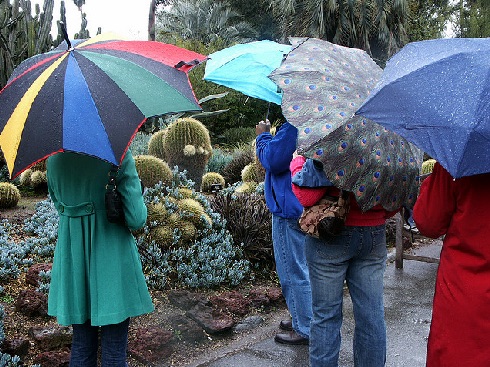}&
    \includegraphics[width=1.21in]{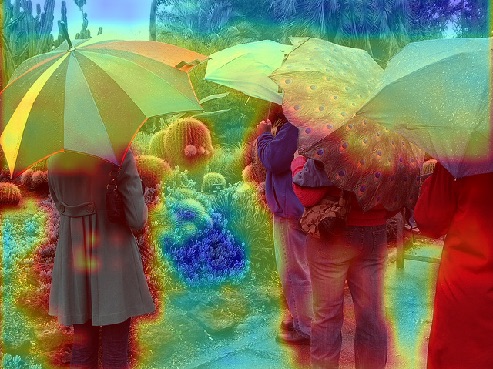}&
    \includegraphics[width=1.21in]{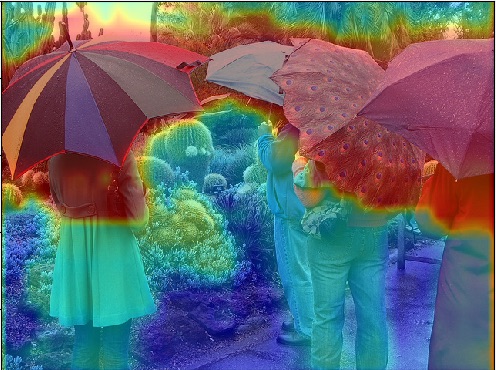}& 
    \includegraphics[width=1.21in]{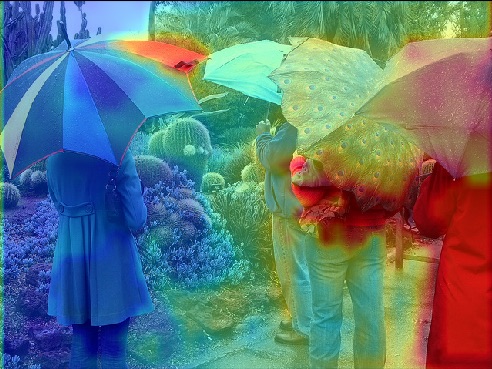}&
     \includegraphics[width=1.21in]{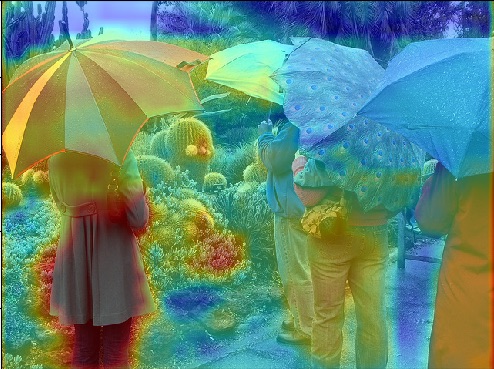}\\
     
      & \textit{``woman"} & \textit{``umbrella"} & \textit{``red"} & \textit{``a woman in}\\  &&&& \textit{a green coat"} \\ 

     \includegraphics[width=1.21in]{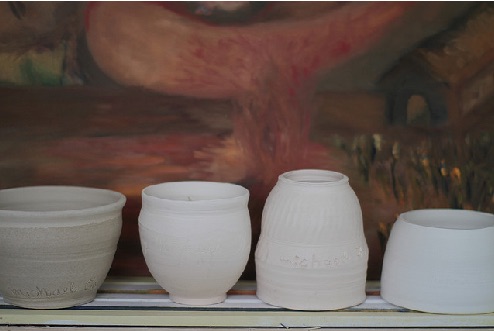}&
    \includegraphics[width=1.21in]{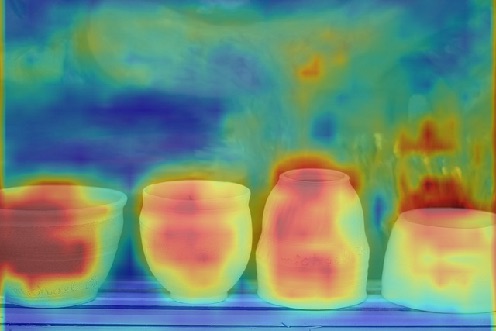}&
    \includegraphics[width=1.21in]{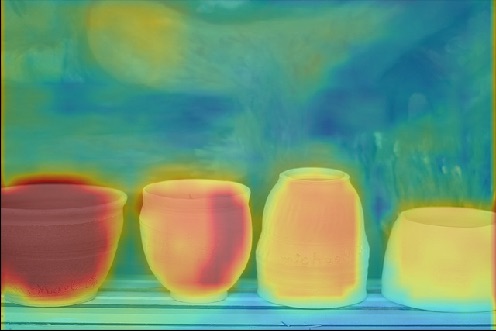}& 
    \includegraphics[width=1.21in]{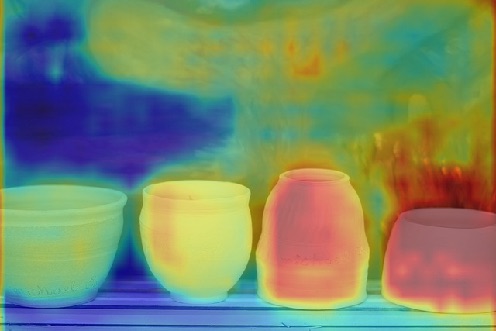}&
     \includegraphics[width=1.21in]{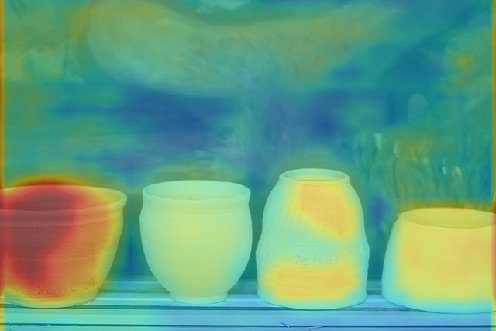}\\
     
      & \textit{``pot"} & \textit{``pot on left"} & \textit{``two pot on the right"} &  \textit{``the pot on the far left"} \\ 
 \end{tabular}
}
}
\caption{(Best viewed in color) Visualization of spatial feature representation. These spatial heatmaps show the responses of the network to different query expressions.}
\label{fig:word}
\end{figure*}

\begin{figure}[htb]
  \centering
{\setlength{\tabcolsep}{0.5pt}
\scalebox{0.84}{
  \begin{tabular}{ccc}

         \multicolumn{3}{c}{Query: \textit{``boy on right''}}    \\
    \includegraphics[width=1.2in]{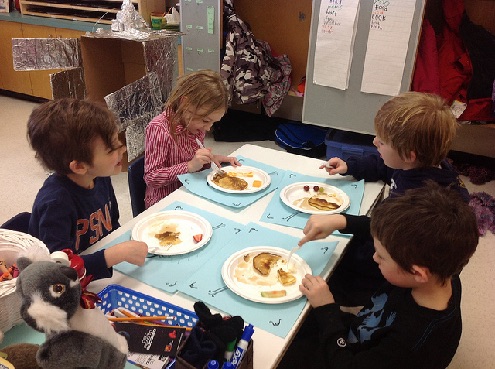}&
    \includegraphics[width=1.2in]{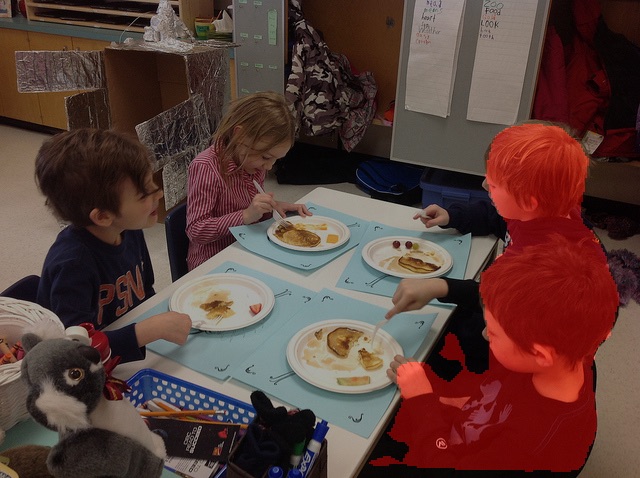}&
    \includegraphics[width=1.2in]{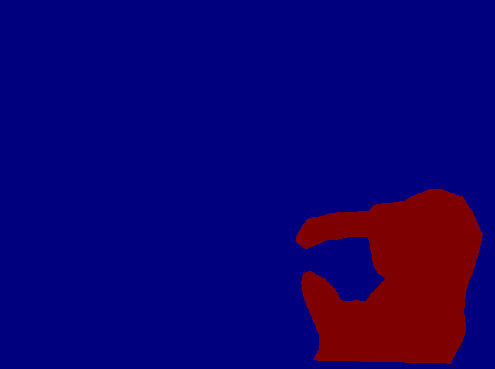}\\
    
      \multicolumn{3}{c}{Query: `` \textit{legs on left''}}    \\
    \includegraphics[width=1.2in]{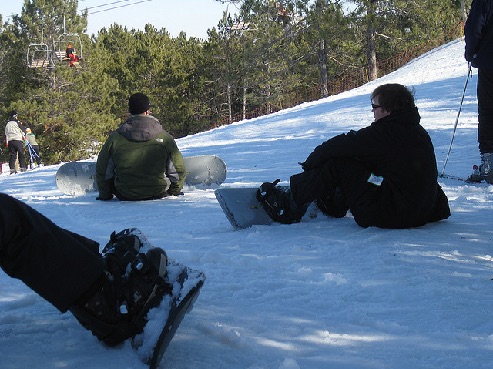}&
    \includegraphics[width=1.2in]{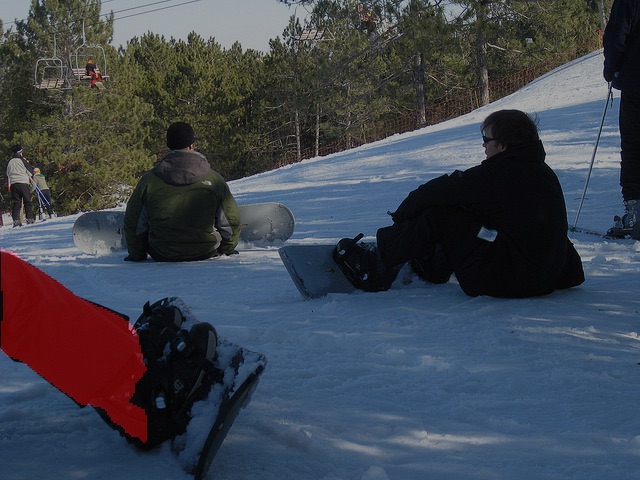}&
    \includegraphics[width=1.2in]{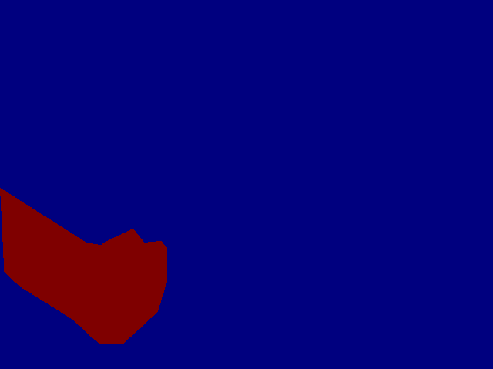}\\
    
    \multicolumn{3}{c}{Query:  \textit{``car next to cab''}}    \\
    \includegraphics[width=1.2in]{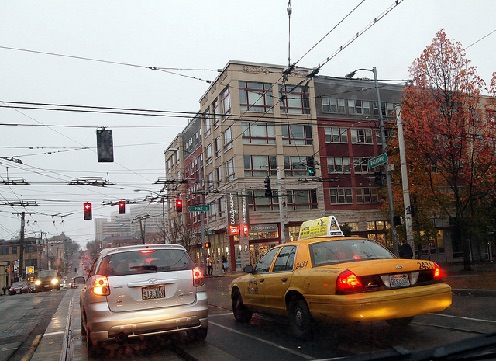}&
    \includegraphics[width=1.2in]{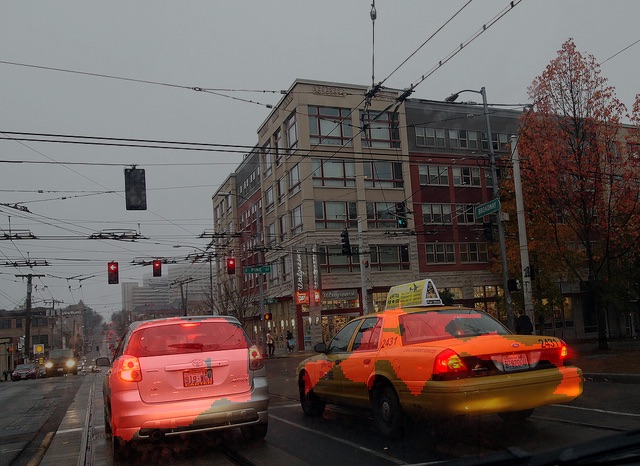}&
    \includegraphics[width=1.2in]{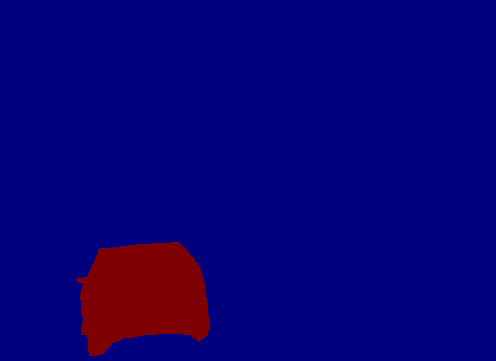}\\

      \multicolumn{3}{c}{Query:  \textit{``right bike''}}    \\  
    \includegraphics[width=1.2in]{figure/Fig_failure/462_im.jpg}&
    \includegraphics[width=1.2in]{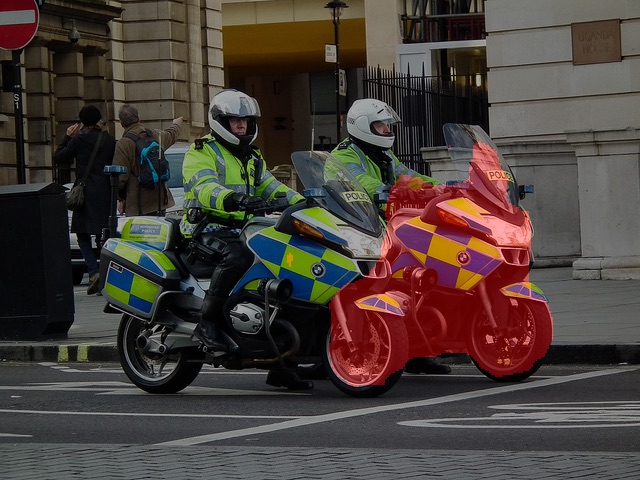}&
    \includegraphics[width=1.2in]{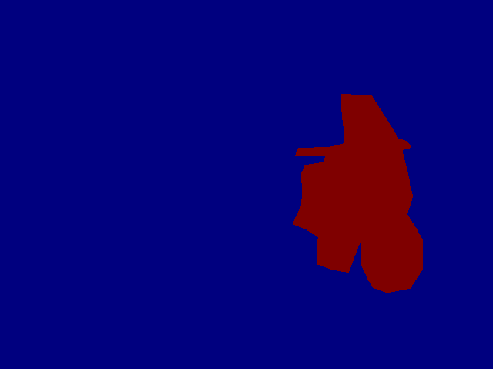}\\

       (a) & (b) & (c) 
  \end{tabular}
}
}
\caption{Some failure examples of our model: (a) original image; (c) segmentation result; (c) ground truth. The failures are due to factors such as language ambiguity (1st and 2nd rows), similar object appearance (3rd row) and occlusion (4th row).}
\label{fig:failure}
\end{figure}

\subsection{Visualization and Failure Cases}\label{sec:vis}

\noindent{\bf Visualization}: We visualize spatial feature representations with various query expressions for a given image. This helps to gain further insights on the learned model. 

We adopt the same technique in~\cite{li2018referring} to generate visualization heatmaps over spatial locations. It is created by normalizing the strongest activated channel of the last feature map, which is upsampled to match with the size of original input image. These generated heatmaps are shown in Fig.~\ref{fig:word}. It can be observed that our model is able to correctly respond to different query expressions with various categories, locations and relationships. For instance, in the second row, when the query is ``woman'' and ``umbrella'', our model highlights every woman and umbrella in the image. Similarly, when the query is ``red'', it captures both the red clothes and the red part of the umbrella. For a more specific phrase such as ``a woman in a green coat'', the model accurately identifies the woman being referred to.

\noindent{\bf Failure cases}: We also visualize some interesting failure cases in Fig.~\ref{fig:failure}. These failures are caused by the ambiguity of the language (e.g. two boys on right in the 1st example and the feet in the 2nd example), similar object appearance (e.g. car vs cab in the 3rd example), and occlusion (the wheel of the motorbike in the 4th example). Some of these failure cases can potentially be fixed by applying object detectors.


\section{Conclusion}
We have proposed cross-modal self-attention and gated multi-level fusion modules to address two crucial challenges in the referring image segmentation task. Our cross-modal self-attention module captures long-range dependencies between visual and linguistic modalities, which results in a better feature representation to focus on important information for referred entities. In addition, the proposed gated multi-level fusion module adaptively integrates features from different levels via learnable gates for each individual level. The proposed network achieves state-of-the-art results on all four benchmark datasets.

\noindent{\bf Acknowledgements}: LY, MR and YW are supported by NSERC. ZL is supported by the National Natural Science Foundation of China under Grant No. 61771301. LY and MR are also supported by the GETS and UMGF programs at the University of Manitoba. Thanks to NVIDIA for donating some of the GPUs used in this work.



{\small
\bibliographystyle{ieee_fullname}
\bibliography{refseg_nl}
}

\end{document}